# Prompt Engineering Large Language Models' Forecasting Capabilities

Philipp Schoenegger (LSE), Cameron Jones (UC San Diego),
Philip E. Tetlock (Wharton), Barbara Mellers (Wharton)

**Abstract**

*Large language model performance can be improved in a large number of ways. Many such techniques, like fine-tuning or advanced tool usage, are time-intensive and expensive. Although prompt engineering is significantly cheaper and often works for simpler tasks, it remains unclear whether prompt engineering suffices for more complex domains like forecasting. Here we show that small prompt modifications rarely boost forecasting accuracy beyond a minimal baseline. In our first study, we tested 38 prompts across Claude 3.5 Sonnet, Claude 3.5 Haiku, GPT-4o, and Llama 3.1 405B. In our second, we introduced compound prompts and prompts from external sources, also including the reasoning models o1 and o1-mini. Our results show that most prompts lead to negligible gains, although references to base rates yield slight benefits. Surprisingly, some strategies showed strong negative effects on accuracy: especially encouraging the model to engage in Bayesian reasoning. These results suggest that, in the context of complex tasks like forecasting, basic prompt refinements alone offer limited gains, implying that more robust or specialized techniques may be required for substantial performance improvements in AI forecasting.*

## 1. Introduction[1]

Forecasting future events has significant decision-relevance, as having a well-calibrated probabilistic estimation of the risk of a future pandemic, a conflict, or an emerging technology is crucial in making decisions under uncertainty. Current best practices for forecasting rely on aggregating the judgemental forecasts of experienced forecasters (Tetlock & Gardner 2016), a process that is both lengthy and expensive, though it promises to produce valuable input into decision-making processes (Mellers et al, 2019; Tetlock et al. 2014).

Recent work has applied frontier large language models (LLM) to forecasting, testing a variety of research questions, such as whether LLMs are able to match human forecasting performance, what determines their prediction capabilities, and how these capabilities may be increased. For example, previous work looked at retrieval-augmented systems (Halawi et al. 2024), aggregation of multiple models (Schoenegger et al. 2024), ranking-based context retrieval systems (Yan et al. 2024), or applications of reinforcement learning (Turtel et al. 2025b). While many of these approaches have resulted in increased forecasting performance, the current performance of frontier models still trails experienced forecaster aggregates on ForecastBench (Karger et al. 2024).

Many such approaches have focused on specific aspects in designing forecasting pipelines such as effective news aggregation (Wang et al. 2025) or fine-tuning on model self-play output (Turtel et al. 2025). However, prompting is a significantly cheaper and easier to implement technique that has not received as much attention as others. Additionally, existing work showing improvements in forecasting performance from modifying other aspects of the pipeline is likely to only use a single prompt or vary a small set of prompts. This is, in part, due to resource constraints as multiplying any design to test for a variety of prompting approaches is often not feasible. However, this threatens leading to reduced overall capabilities as comparatively low-effort changes in prompting are not investigated sufficiently, suggesting that low-cost improvements to many LLM forecasting approaches might yet be found. This is in part because systematic reviews of prompt engineering efforts in related topics are relatively sparse, making efficient prompt selection relatively difficult.

---

[1] This research was supported by a grant from the Wharton AI & Analytics Initiative.

Prompting, using natural language instructions to improve instruction following and performance more generally, has been a cornerstone of LLM research since GPT-2, which showed strong translation capabilities via a structured prompt that helped the model understand the task (Radford et al. 2019). The next level of prompt design was documented in GPT-3, by showing that few-shot approaches via text interaction lead to increased performance (Brown et al. 2020). Over the past few years, "prompt engineering" has blossomed into a quickly-maturing field, with research ranging from automated prompt generation (Shin et al. 2020) and selection (Do et al. 2024) to theory-driven approaches such as EchoPrompt (Mekala et al. 2023), the now-standard chain-of-thought prompting (Wei et al. 2022) that raised reasoning capabilities across the board by asking the model to think step-by-step through a problem, or simply using emotional appeals in the prompt (Li et al. 2023).

Prompt practices and usage varies widely by field, with requirements for prompt engineering in cyber defense (Shenoy & Mbaziira 2024) differing from prompt approaches in software engineering (Shin et al. 2023), as well as by the type and size of model used (He et al. 2024; Ma et al. 2024). Some sets of prompts directly draw on the model architecture, such as by increasing inference time compute by restating questions or allowing the model to reason through more complex problems via a chain of thought. Other prompts more directly draw on human psychology, such as emotional stimuli. This has led to a large number of prompting approaches and systematisation attempts (Schulhoff et a. 2024; Sahoo et al. 2024). However, there has been no broad effort to test and systematise prompts in the context of forecasting. As prompting often directly depends on the specific field that the models are being deployed in, this suggests that this may be a fruitful task in attempting to understand and improve model forecasting capabilities. We also hope that our results can be directly adapted by other LLM forecasting researchers, as changes in prompts are often very low-cost changes that can be made to many forecasting systems with minimal effort.

As such, we begin this paper by testing a large base of 38 distinct prompts to improve our understanding of the effect of prompts on forecasting performance of LLMs, ranging from standard prompts motivated by model architecture to more domain-specific and human psychology derived prompts. We first compare 37 treatment prompts to a control across various models (Claude 3.5 Sonnet, GPT-4o, Claude 3.5 Haiku, Llama 3.1 405B), finding that most prompts do not significantly alter forecasting performance, though some (base rate usage) increase performance. Surprisingly, we also do find that some prompts(Bayesian reasoning) decrease accuracy. In a second study, we use these results to construct composite prompts and compare them to automated prompts from major model developers and prompts used in the literature on the same set of models as well as reasoning models (o1 and o1-mini), finding similar results: no prompt robustly improved performance. We suggest that in the context of forecasting, prompt engineering is unlikely to play a major role in improving model performance.

## 2. Study 1
### 2.1. Method
For Study 1, we test the effectiveness of 37 distinct prompts (with respect to a control prompt) on the forecasting accuracy of 4 LLMs at a question set of 100 forecasting questions. We preregistered our study procedure and analysis plan at the Open Science Framework[2].

#### 2.1.1. Models
We use a total of four non-reasoning models. These include standard frontier models (Claude 3.5 Sonnet, GPT-4o), small models (Claude 3.5 Haiku), as well as open weight models (Llama 3.1 405B). Specifically, we use the following API endpoints via AWS Bedrock: Llama 3.1 405B (meta.llama3-1-405b-instruct-v1:0), Claude 3.5 Sonnet (anthropic.claude-3-5-sonnet-20241022-v2:0), and Claude 3.5 Haiku (anthropic.claude-3-5-haiku-20241022-v1:0). We queried all models at a temperature of 0 to reduce variation in outputs and set a maximum of 2000 output tokens. We did not re-run any queries if the model refused to respond for any reason.

All models have a training data cut-off of July 2024 or before, with Claude 3.5 Haiku having a cut-off date of July 2024, Claude 3.5 Sonnet of July 2024, Llama 3.1 405B of December 2023, and GPT-4o and o1-mini of October 2023. This allows us to use forecasting questions that resolve at the end of 2024, as no model can have the answers as part of their training data. Importantly, the fact that the training data cut-off dates vary by up to nine months does not matter for our design as we do not directly compare accuracy between models but instead look at differences between prompts across models.

#### 2.1.2. Prompts
We designed a total of 38 prompts, which include 1 control prompt and 37 treatment prompts. In total, three topic areas inspired our choice of prompts. First, we selected a large number of prompts that have been empirically established in the literature on LLMs, such as chain of thought reasoning (Wei et al. 2022) or expert prompting (Kong et al. 2023). Second, we also directly drew on the existing social science literature on forecasting, building upon results from human studies, such as usage of base rates or abstaining from forecasting difficult questions. Third, we used a set of prompts that were inspired by the general literature on human judgement and decision-making, such as anti-biasing (Furnham & Boo 2011) and metacognition (Flavell 1979).

Our set of prompts draw on a variety of prompt types that we formulated prior to conducting the study. We divide our prompts into the following types: (i) Unguided Reflection, (ii) Framework, (iii) Information, and (iv) Incentives. Prompts of the Unguided Reflection type prompt the model to think or reflect without providing a specific structure or method to this reflection. In contrast, the prompts of the Framework type set up specific formats, methods, or approaches to generating the forecast. Prompts of the Information type, on the other hand, provide or request additional information, whereas Incentive-type prompts provide motivation or stakes to encourage more accurate forecasting.

---
[2] https://osf.io/ncph7/?view_only=a19c94732136469d83ddbefe7ed837b1

We also divide our prompts by cognitive mode prior to data collection. Specifically, we divide our prompts as roughly following the following cognitive modes: (i) Analytical, (ii) Intuitive, (iii) Reference-Based, and (iv) Dialectical. Analytical prompts are systematic, encourage explicit reasoning, focus on decomposition and favor clear steps. On the other hand, intuitive prompts focus on gut feelings, pattern recognition, and provide a more holistic judgement. Reference-based prompts primarily draw on external comparisons and benchmarks and focus on similar cases or base rates. Lastly, dialectical prompts include explicit consideration of opposing views and focus on synthesis of contradictions.

We compare all prompts against the following control prompt.

> *Please answer the following question with a probabilistic estimate expressed between 0% and 100%, and format your response as: 'Forecast: X%'.*
> *{Question}*

Our treatment prompts add or otherwise modify this base control prompt. For example, the chain of thought (CoT) treatment prompt adds 'Approach this question step-by-step, explaining your reasoning at each stage' to the control prompt. This allows us to identify the effect of a single change in prompting on final forecasting accuracy. Below we outline a sample of our prompts, see Table 1. For a full list of all prompts, see Appendix A.

**Table 1. Example Prompts (Study 1)**

| Prompt Name | Source/ Motivation | Prompt Text | Prompt Type | Cognitive Mode |
|---|---|---|---|---|
| Echo | Mekala et al. (2023) | Please answer the following question. First, let's repeat the question, and then provide a probabilistic estimate expressed between 0% and 100%, and format your response as: 'Forecast: X%'.<br><br>{Question} | Unguided Reflection | Intuitive |
| Base Rate First | Tetlock & Gardner (2016) | Please answer the following question with a probabilistic estimate expressed between 0% and 100%, and format your response as: 'Forecast: X%'.<br><br>Before considering the specific details of this question, what is the historical frequency of similar events? Using this base rate as your starting point, | Information | Reference-based |

| | | adjust your probability estimate based on the particular circumstances of this case.<br><br>{Question} | | |
|---|---|---|---|---|
| Anti-Biasing (Overconfidence). | Fellner & Krügel (2012). | Please answer the following question with a probabilistic estimate expressed between 0% and 100%, and format your response as: 'Forecast: X%'. Be cautious of overconfidence by carefully considering uncertainties in your prediction.<br><br>{Question} | Framework | Analytical |

Notes. *The table lists some of the prompts used in Study 1 as well as their prompt type and cognitive mode.*

### 2.1.3. Question Set

We use questions drawn from the question set of the forecasting benchmark ForecastBench (Karger et al. 2024). Specifically, we sample 100 questions from their 2024-12-08 data set, which has questions resolving in December 2024, months after the most recent training data cutoff of the included models. The questions we sampled were from the following sources: DBnomics, FRED, INFER, Manifold Markets, Metaculus, Polymarket, Wikipedia, and Yahoo Finance. All questions are binary, i.e., they resolve either as 'Yes' or 'No'. The questions included topics like conflict, economics, weather, elections, popular culture, and technology. For a sample of the questions used, see Table 2.

**Table 2. Sample ForecastBench Questions Included in the 100-Item Test Set**

| Source | Question |
|---|---|
| Manifold Markets | Will Ukraine regain control over Crimea before the end of 2024? |
| FRED | Will the 90-day average of the Federal Reserve's Secured Overnight Financing Rate have increased by 2024-12-15 as compared to its value on 2024-12-08? |
| DBnomics | What is the probability that the daily average temperature at the French weather station at Clermont-Ferrand Auvergne Airport will be higher on 2024-12-15 than on 2024-12-08? |
| Metaculus | Will X declare bankruptcy in 2024? |
| Polymarket | Will four SpaceX Starship launches successfully reach outer space in 2024? |
| Wikipedia | [According to Wikipedia] Will a vaccine have been developed for Fusobacterium infection by 2024-12-15? |

Notes. *Example questions used in both Study 1 and Study 2. Drawn from ForecastBench.*

## 2.2. Results

We aimed to collect a total of 3,800 forecasts from each model. GPT-4o had the highest refusal rate (4.32%), followed by Haiku (3.16%) and Sonnet (2.79%), while Llama 3.1 405B had the lowest refusal rate (0.76%). In total, 419 forecasts were missing across all models. 38% of total refusals occurred in the Abstention prompt, which instructed the models to not give a forecast if they believe that they cannot forecast the question. After excluding all missing forecasts, our final data set consists of 3636 forecasts for GPT-4o, 3680 for Claude 3.5 Haiku, 3771 for Llama 3.1 405B, and 3694 for Claude 3.5 Sonnet.

While most forecasts had mean forecasts somewhat close to the 50% midpoint, we do find some variation in how much prompts led models to deviate from the midpoint, with Bayesian Reasoning, Propose-Evaluate-Select, and Chain of Thought prompts making more decisive forecasts, with mean differences between 17.2 and 19.9 percentage points. Conversely, some prompts led to forecasts that were, on average, closer to the midpoint. This was the case for Anti-Biasing (Anchoring), Re-Reading, and Control prompts, which differed only between 11.9 and 13.4 percentage points from the midpoint on average, see Figure 1. Across prompts, forecasts were also less likely to be close to the midpoint on questions from Metaculus, Wikipedia, and Polymarket, while FRED, YahooFinance, and DBnomics resulted in forecasts that were closer to the middle.

**Figure 1. Distribution of Forecasts by Model.**

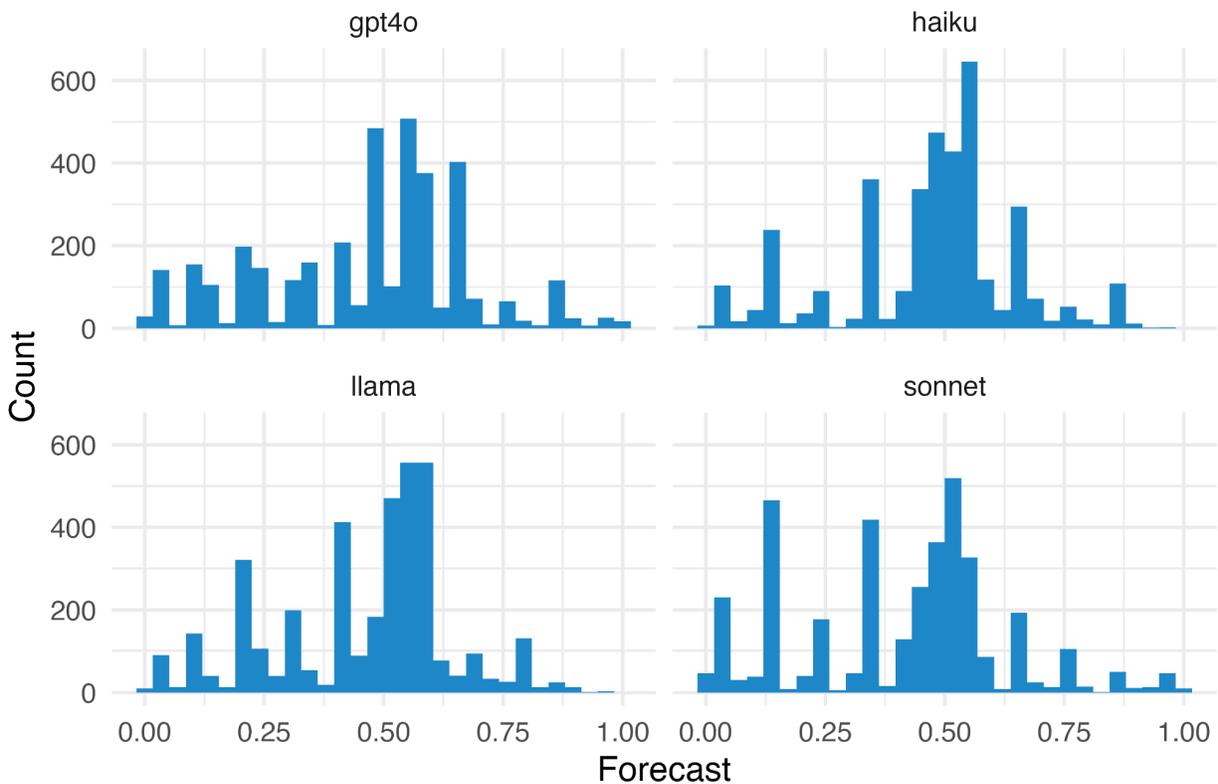

Notes. *Modal and median forecasts were close to 0.5, indicating models expressed uncertainty about many questions and that predictions were relatively balanced with respect to outcome. Nevertheless, some predictions were more decisive (closer to 0 or 1). There was little variation in forecast distribution by model.*

**Table 3. Mean Forecasts**

| Prompt Name | Mean Forecast (SD) | Mean Absolute Distance |
|---|---|---|
| Bayesian Reasoning | 0.476 (0.250) | 0.199 |
| Propose-Evaluate-Select | 0.470 (0.224) | 0.175 |
| Chain of Thought | 0.435 (0.221) | 0.172 |
| Fermi Estimate | 0.456 (0.220) | 0.172 |
| Superforecasting Persona (Short) | 0.453 (0.217) | 0.171 |
| Event Decomposition | 0.460 (0.207) | 0.167 |
| Self-critique | 0.429 (0.223) | 0.165 |
| Question Paraphrasing | 0.433 (0.210) | 0.165 |
| Self-Consistency | 0.479 (0.205) | 0.163 |
| Time Decomposition | 0.437 (0.204) | 0.162 |
| Counterfactual Reasoning | 0.459 (0.212) | 0.162 |
| Tipping | 0.458 (0.213) | 0.160 |
| Analogical Reasoning | 0.458 (0.203) | 0.160 |
| Base Rate First | 0.428 (0.207) | 0.159 |
| Uncertainty Quantification | 0.453 (0.203) | 0.156 |
| Few-Shot | 0.472 (0.193) | 0.154 |
| Premortem | 0.469 (0.196) | 0.154 |
| Pros & Cons | 0.448 (0.191) | 0.152 |
| Frequency-Based Reasoning | 0.427 (0.202) | 0.152 |
| Step-Back | 0.455 (0.200) | 0.151 |
| Metacognition | 0.465 (0.195) | 0.149 |
| Scoring Rule | 0.453 (0.191) | 0.148 |
| Abstention | 0.449 (0.193) | 0.147 |
| Hypothetical Scenario Analysis | 0.444 (0.192) | 0.147 |
| Deep Breath | 0.447 (0.190) | 0.146 |
| Structure | 0.451 (0.189) | 0.143 |
| Explicit Uncertainty Sources | 0.458 (0.183) | 0.142 |
| High Personal Stakes | 0.451 (0.191) | 0.142 |
| Anti-Biasing (Round Numbers) | 0.479 (0.183) | 0.141 |
| Multiple Reference Classes | 0.443 (0.184) | 0.139 |
| Simulated Dialogue | 0.460 (0.170) | 0.139 |
| Simulated Debate | 0.462 (0.173) | 0.139 |
| Echo | 0.446 (0.187) | 0.138 |
| Anti-Biasing (Overconfidence) | 0.458 (0.180) | 0.137 |
| Emotional Prompt | 0.461 (0.183) | 0.134 |
| Control | 0.453 (0.185) | 0.134 |
| Re-Reading | 0.444 (0.180) | 0.133 |
| Anti-Biasing (Anchoring) | 0.441 (0.172) | 0.119 |

Notes. *Mean forecast with standard deviation in parentheses as well as mean of absolute distance to the 50% midpoint. Ordered by mean absolute distance.*

We first calculate accuracy of forecasts via Brier scores, where lower scores indicate higher accuracy. Formally, the Brier score is defined as $BS = (1/N) * \Sigma (p_i - o_i)^2$ for $i = 1$ to $N$, where $p_i$ is the predicted probability for outcome i and $o_i$ is the actual (observed) outcome (0 or 1). A lower Brier score reflects forecasts that better match the observed outcomes. Intuitively, it measures how close the probability estimates are to the eventual binary results. While between-model comparisons of accuracy are not the purpose of this study, we find some variation in accuracy (mean Brier scores, with standard deviations in parentheses): GPT-4o, 0.215 (0.161); Claude 3.5 Haiku, 0.214 (0.132); Llama 3.1 405B, 0.227 (0.159); and Claude 3.5 Sonnet, 0.208 (0.168).

To test the effects of prompts compared to the control across all questions and models, we computed a difference score for each forecast by subtracting the control prompt's Brier score (for the same question and model) from the experimental prompt's Brier score. We then fit a preregistered linear mixed-effects model predicting this difference, with each prompt type as a fixed effect and random intercepts for questions and models. This setup accounts for variability in question difficulty and baseline model performance. Finally, we used estimated marginal means to test whether each prompt's average difference from control was significantly different from zero, applying Benjamini-Hochberg corrections for multiple comparisons, see Table 4.

**Table 4. Mixed-Effects Estimates of Prompt Impact on Forecasting Accuracy (Study 1)**

| Prompt Name | Estimate | Z-ratio | p-value | Adj. p |
|---|---|---|---|---|
| Frequency-Based Reasoning | -0.014 | -2.563 | 0.128 | 0.774 |
| Base Rate First | -0.011 | -2.069 | 0.253 | 0.774 |
| Chain Of Thought | -0.011 | -2.043 | 0.253 | 0.774 |
| Step-Back | -0.011 | -1.974 | 0.256 | 0.774 |
| Pros & Cons | -0.009 | -1.623 | 0.387 | 0.774 |
| Premortem | -0.008 | -1.468 | 0.4 | 0.774 |
| Question Paraphrasing | -0.008 | -1.461 | 0.4 | 0.774 |
| Time Decomposition | -0.008 | -1.434 | 0.4 | 0.774 |
| Hypothetical Scenario Analysis | -0.007 | -1.347 | 0.411 | 0.774 |
| Echo | -0.006 | -1.118 | 0.488 | 0.774 |
| Uncertainty Quantification | -0.006 | -1.021 | 0.541 | 0.774 |
| Explicit Uncertainty Sources | -0.005 | -0.927 | 0.565 | 0.774 |
| Metacognition | -0.005 | -0.9 | 0.565 | 0.774 |
| Deep Breath | -0.005 | -0.877 | 0.565 | 0.774 |
| Re-Reading | -0.005 | -0.836 | 0.565 | 0.774 |
| Counterfactual Reasoning | -0.004 | -0.826 | 0.565 | 0.774 |
| Multiple Reference Classes | -0.004 | -0.82 | 0.565 | 0.774 |
| Self-Critique | -0.004 | -0.64 | 0.644 | 0.794 |
| Simulated Debate | -0.003 | -0.579 | 0.671 | 0.801 |
| Analogical Reasoning | -0.002 | -0.386 | 0.809 | 0.935 |
| Superforecasting Persona (Short) | -0.001 | -0.213 | 0.932 | 0.997 |
| Structure | 0.0 | 0.004 | 0.997 | 0.997 |
| Abstention | 0.0 | 0.025 | 0.997 | 0.997 |
| Anti-biasing (overconfidence). | 0.0 | 0.069 | 0.997 | 0.997 |
| Anti-biasing (Anchoring) | 0.001 | 0.091 | 0.997 | 0.997 |
| Emotional Prompt | 0.004 | 0.66 | 0.644 | 0.794 |
| High Personal Stakes | 0.004 | 0.717 | 0.626 | 0.794 |
| Tipping | 0.006 | 1.151 | 0.486 | 0.774 |
| Scoring Rule | 0.006 | 1.176 | 0.486 | 0.774 |
| Event Decomposition | 0.007 | 1.236 | 0.471 | 0.774 |
| Few-Shot | 0.007 | 1.357 | 0.411 | 0.774 |
| Fermi Estimate | 0.008 | 1.548 | 0.4 | 0.774 |
| Self-Consistency | 0.009 | 1.717 | 0.361 | 0.774 |
| Simulated Dialogue | 0.009 | 1.707 | 0.361 | 0.774 |
| Anti-Biasing (Round Numbers) | 0.011 | 2.097 | 0.253 | 0.774 |
| Bayesian Reasoning | 0.03 | 5.436 | **<0.001** | **<0.001** |
| Propose-Evaluate-Select | 0.033 | 6.118 | **<0.001** | **<0.001** |

Notes. *Linear mixed-effects model results. Prompt type is a fixed effect, random intercepts for questions and models. Negative coefficients indicate higher accuracy than control, positive coefficients indicate worse performance. P-value adjustment is done via the Benjamini-Hochberg procedure. Both before and after adjustment, only Bayesian Reasoning and Propose-Evaluate-Select prompts differ significantly from the control, leading to less accurate forecasts.*

The results indicate that in this model, only the Bayesian Reasoning and Propose-Evaluate-Select prompts show statistically significant effects compared to the control. Importantly, both prompts lead to less accurate forecasts than the control. No other prompt differs significantly from the control.

**Figure 2. Prompt-Level Accuracy Changes Relative to Control (Study 1)**

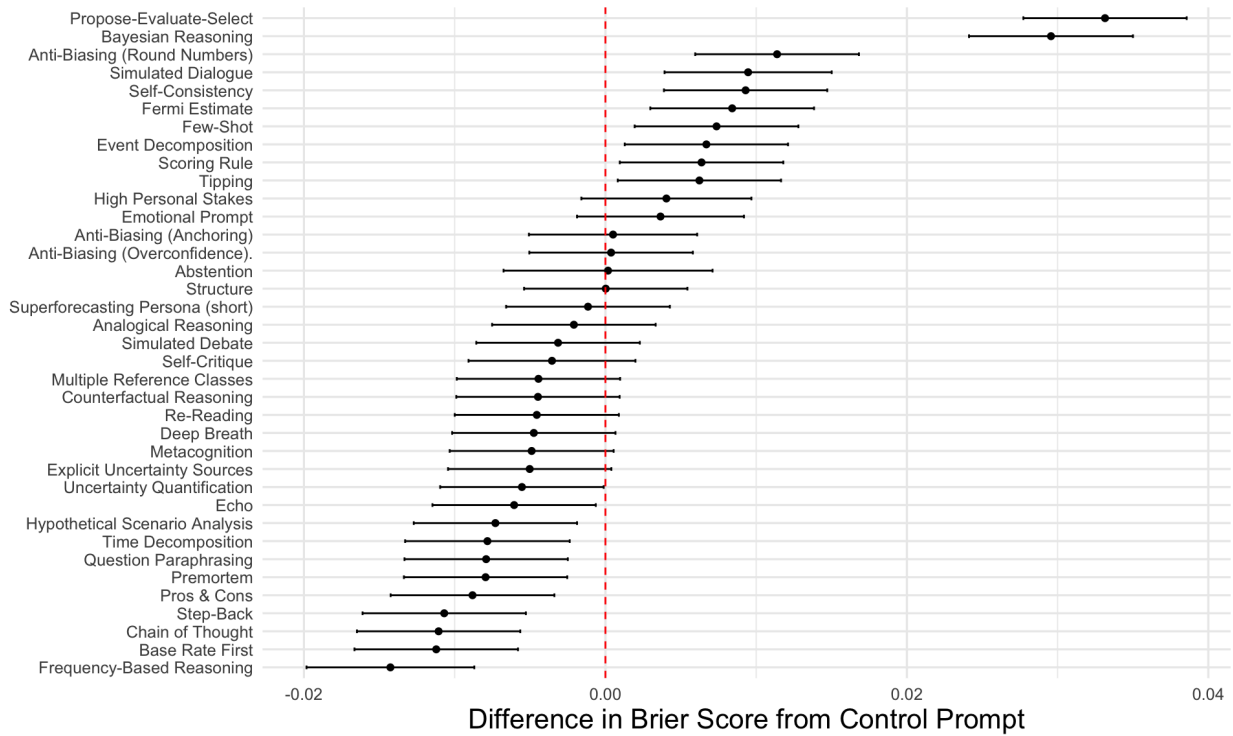

Notes. *Each point shows the estimated difference from the control prompt in the linear mixed-effects model with random intercepts for questions and models. Negative values on the y-axis indicate better performance (i.e., lower Brier scores) compared to the control prompt, whereas positive values indicate worse performance. Error bars represent ±1 standard error. The dashed horizontal line at zero indicates no difference from the control condition.*

We also preregistered a simpler analysis that ignores question- and model-level clustering and instead runs one-sample t-tests comparing each prompt's difference-from-control against zero, see Table 5.

**Table 5. One-Sample t-Tests Results of Prompt Impact on Forecasting Accuracy (Study 1)**

| Prompt Name | Mean Diff (SD) | N | t-statistic | p-value | Adj p |
| --- | --- | --- | --- | --- | --- |
| Frequency-Based Reasoning | -0.019 (0.114) | 370 | -3.267 | **0.001** | **0.022** |
| Base Rate First | -0.016 (0.106) | 390 | -2.971 | **0.003** | **0.036** |
| Chain Of Thought | -0.016 (0.128) | 391 | -2.431 | **0.015** | 0.086 |
| Step-Back | -0.015 (0.107) | 391 | -2.835 | **0.005** | **0.036** |
| Pros & Cons | -0.014 (0.111) | 389 | -2.415 | **0.016** | 0.086 |
| Time Decomposition | -0.013 (0.149) | 385 | -1.667 | 0.096 | 0.210 |
| Premortem | -0.013 (0.106) | 391 | -2.351 | **0.019** | 0.089 |
| Question Paraphrasing | -0.013 (0.127) | 391 | -1.960 | 0.051 | 0.146 |
| Hypothetical Scenario Analysis | -0.012 (0.107) | 391 | -2.213 | **0.027** | 0.113 |
| Echo | -0.011 (0.097) | 391 | -2.172 | **0.030** | 0.113 |
| Uncertainty Quantification | -0.010 (0.116) | 390 | -1.755 | 0.080 | 0.185 |
| Explicit Uncertainty Sources | -0.010 (0.102) | 391 | -1.879 | 0.061 | 0.151 |
| Metacognition | -0.010 (0.099) | 388 | -1.922 | 0.055 | 0.146 |
| Self-Critique | -0.009 (0.141) | 374 | -1.294 | 0.197 | 0.346 |
| Deep Breath | -0.009 (0.092) | 391 | -2.022 | 0.044 | 0.146 |
| Counterfactual Reasoning | -0.009 (0.126) | 391 | -1.427 | 0.154 | 0.301 |
| Multiple Reference Classes | -0.009 (0.113) | 391 | -1.588 | 0.113 | 0.233 |
| Re-Reading | -0.009 (0.092) | 387 | -1.941 | 0.053 | 0.146 |
| Simulated Debate | -0.008 (0.147) | 390 | -1.052 | 0.293 | 0.428 |
| Analogical Reasoning | -0.007 (0.126) | 390 | -1.074 | 0.284 | 0.428 |
| Superforecasting Persona (Short) | -0.006 (0.114) | 389 | -1.036 | 0.301 | 0.428 |
| Structure | -0.005 (0.101) | 391 | -0.902 | 0.368 | 0.504 |
| Anti-Biasing (Anchoring) | -0.005 (0.076) | 368 | -1.153 | 0.250 | 0.402 |
| Anti-Biasing (Overconfidence) | -0.004 (0.070) | 390 | -1.241 | 0.215 | 0.362 |
| Abstention | -0.002 (0.089) | 238 | -0.332 | 0.740 | 0.818 |
| High Personal Stakes | 0.000 (0.092) | 360 | -0.084 | 0.933 | 0.949 |
| Emotional Prompt | 0.000 (0.091) | 374 | -0.064 | 0.949 | 0.949 |
| Tipping | 0.002 (0.137) | 391 | 0.229 | 0.819 | 0.866 |
| Scoring Rule | 0.002 (0.097) | 390 | 0.339 | 0.734 | 0.818 |
| Event Decomposition | 0.002 (0.128) | 391 | 0.317 | 0.751 | 0.818 |
| Few-Shot | 0.003 (0.110) | 389 | 0.471 | 0.638 | 0.767 |
| Fermi Estimate | 0.004 (0.153) | 389 | 0.464 | 0.643 | 0.767 |
| Simulated Dialogue | 0.004 (0.135) | 373 | 0.599 | 0.549 | 0.701 |
| Self-Consistency | 0.005 (0.139) | 391 | 0.663 | 0.507 | 0.671 |
| Anti-Biasing (Round Numbers) | 0.007 (0.092) | 389 | 1.398 | 0.163 | 0.301 |
| Bayesian Reasoning | 0.025 (0.172) | 388 | 2.854 | **0.005** | **0.036** |
| Propose-Evaluate-Select | 0.028 (0.153) | 391 | 3.685 | **0.000** | **0.010** |

Notes. *One-sample t-test results. Negative values of the mean difference indicate higher accuracy than control, while positive values indicate worse performance. P-values are adjusted for multiple comparisons using the Benjamini-Hochberg procedure. After correction, three prompts (Frequency-Based Reasoning, Base Rate First, and*

*Step-Back) significantly outperformed the control, whereas two prompts (Bayesian Reasoning and Propose-Evaluate-Select) performed significantly worse. All remaining prompts did not differ significantly from the control after adjustment.*

We find that eight prompts showed significantly better performance than the control at $p < .05$ prior to p-value adjustment (Frequency-Based Reasoning, Base Rate First, Step-Back, Chain of Thought, Pros & Cons, Premortem, Hypothetical Scenario Analysis, and Echo), while two prompts (Bayesian Reasoning and Propose-Evaluate-Select) performed worse. After applying Benjamini–Hochberg corrections, only three prompts remained significantly better (Frequency-Based Reasoning, Base Rate First, and Step-Back), and two prompts remained significantly worse (Bayesian Reasoning and Propose-Evaluate-Select).

Among prompts identified as significantly improving accuracy (lower Brier scores) relative to the control prompt in the simpler, pre-adjustment one-sample *t*-tests, there was notable diversity across Prompt Types, including both "Information" (Frequency-Based Reasoning, Base Rate First, Pros & Cons) and "Framework" (Step-Back, Chain of Thought, Premortem, Hypothetical Scenario Analysis), as well as variation in Cognitive Modes (Reference-based, Analytical, Dialectical, and Intuitive). Conversely, both prompts consistently identified as significantly worse (Bayesian Reasoning and Propose-Evaluate-Select) varied notably in Prompt Type ("Information" and "Framework," respectively) and Cognitive Mode ("Analytical" and "Dialectical," respectively). Thus, while no single Prompt Type or Cognitive Mode uniformly explains prompt effectiveness, prompts utilizing Reference-based reasoning appear particularly promising see Table 6.

**Table 6. Summary of Significant Prompt Effects Across Prompt Types and Cognitive Modes**

| Prompt Name | Prompt Type | Cognitive Mode | T-test (unadj.) | T-test (BH adj.) | LMM (BH adj.) |
|---|---|---|---|---|---|
| Chain of Thought | Framework | Analytical | Better | NS | NS |
| Step-Back | Framework | Analytical | Better | Better | NS |
| Echo | Unguided Reflection | Intuitive | Better | NS | NS |
| Hypothetical Scenario Analysis | Framework | Dialectical | Better | NS | NS |
| Premortem | Framework | Dialectical | Better | NS | NS |
| Base Rate First | Information | Reference-based | Better | Better | NS |
| Frequency-Based Reasoning | Information | Reference-based | Better | Better | NS |
| Propose-Evaluate-Select | Framework | Dialectical | Worse | Worse | Worse |
| Bayesian Reasoning | Information | Analytical | Worse | Worse | Worse |
| Pros & Cons | Information | Dialectical | Better | NS | NS |
| Deep Breath | Incentives | Intuitive | Better | NS | NS |

Notes. *This table lists each prompt by its Prompt Type (e.g., "Information," "Framework," "Unguided Reflection," "Incentives") and Cognitive Mode (e.g., "Analytical," "Reference-based," "Dialectical," "Intuitive"). It then indicates whether that prompt performed significantly "Better," "Worse," or showed "NS" (no significant difference) compared to the control across three analyses: an unadjusted t‑test, a Benjamini–Hochberg–adjusted t‑test, and a linear mixed‑effects model with Benjamini–Hochberg corrections.*

## 2.3. Discussion

In Study 1, our aim was to test the effect of small changes in prompts, failing to find robust evidence of prompts that improve forecasting accuracy of models, and identifying a few prompts that reduced accuracy. However, these prompts were quite simple and followed only a single idea/theme each. Prompts that are deployed in real-world applications are much more complex and will include multiple steps and elements. In Study 2, we follow up on Study 1 by testing the efficacy of such compound prompts.

## 3. Study 2
### 3.1. Method

For Study 2, we construct a set of compound prompts based on Study 1, and test them alongside automatic prompts from OpenAI and Anthropic, as well as slightly adapted prompts from the LLM forecasting literature. We test this on a total of six models on the same set of 100 forecasting questions from Study 1.

#### 3.1.1. Models

We used the same set of models from Study 1, specifically Claude 3.5 Sonnet, GPT-4o, Claude 3.5 Haiku, Llama 3.1 405B. Additionally, we also run the same for o1 (o1-2024-12-17) and o1-mini (o1-mini-2024-09-12) at a temperature of 1 and with twice the total output tokens at 4000. o1 and o1-mini are of particular interest as so-called "reasoning" models, which are trained to output tokens that are hidden from the user to generate context that can be used to decompose more complex questions. This technique incorporates the key insights from "chain-of-thought" prompting, and so it is of particular interest whether the effectiveness of any prompting techniques generalize to these types of models.

As before, all models had knowledge cutoff dates prior to the question resolutions in December 2024. The new models added in Study 2, o1 and o1-mini have a knowledge cut-off in October 2023, in line with other models from OpenAI used in this study. As before, variation in knowledge cutoff between models is not directly relevant as we do not make between-model accuracy comparisons.

#### 3.1.2. Prompts

We test a total of three sets of prompts against the same control from Study 1. First, we test combinations of successful prompts from Study 1. Second, we also compare automatic prompts constructed via OpenAI's and Anthropic's freely available prompt-design services. Third, we draw on a set of recent academic papers studying LLM forecasting performance, adapting their prompts for our context.

In Study 1, our primary preregistered analysis did not detect any positive effects on any prompts. However, our secondary preregistered measure, while less robust, did show a number of small effects. In order to further test the effects of larger and more complicated prompts, we use the results from this secondary analysis. The prompts that showed positive effects are Chain of Thought, Step-Back, Echo, Hypothetical Scenario Analysis, Premortem, Base Rate First, Frequency-Based Reasoning, Pros & Cons, and Deep Breath. Although we did not identify clear patterns linking these prompts to specific prompt types or cognitive modes, we categorized them based on their originating literature, noting that prompts such as Chain of Thought originated from LLM-focused research, whereas Base Rate First and Frequency-Based Reasoning draw upon human behavioral science and judgmental forecasting literature. Building on these insights, Study 2 systematically examines combinations of these prompts. Specifically, we create pairwise combinations by pairing some prompts from LLM-focused literature with some of those from human behavioral science literature, as well as pairing some prompts exclusively within each literature category. Additionally, we explore the effects of combining all LLM-motivated prompts together, all human-motivated prompts together, and a comprehensive set containing all identified prompts. We make some adjustments to the ordering of phrases to ensure the prompts do not repeat and otherwise result in unintended structured outputs. For the complete list of individual prompts, see Appendix B.

We also use OpenAI's (OpenAI 2025) and Anthropic's (Anthropic 2025) automatic prompt generator as well as Anthropic's automated prompt improver (Anthropic 2024). This allows us to test whether prompt engineering for this context can be fully automated. Lastly, we also collect a series of prompts from published academic work on LLM forecasting. These range from superforecaster persona prompts (Schoenegger et al. 2024), scratchpad prompts (Halawi et al. 2024), and prompts authored by experienced human forecasters (Karger et al. 2024). Testing previously used prompts that are often combinations of many of the individual prompt elements that we have studied provides us with a larger set of combined prompts that may be especially conducive to accuracy. We made small changes to previously used prompts, such as ensuring that the output is in the same style as our other prompts, i.e., 'Forecast: X%'. Additionally, we moved the question text to the bottom of every prompt to ensure the position is held constant across prompts. For a set of example prompts for Study 2, see Table 7.

**Table 7. Example Prompts (Study 2)**

| Prompt Name | Source/Motivation | Prompt Text |
|---|---|---|
| Base Rate First + Chain of Thought | Combined prompt of well-performing prompts drawn from the forecasting and LLM literatures. | Approach this question step-by-step, explaining your reasoning at each stage.<br><br>Before considering the specific details of this question, what is the historical frequency of similar events? Using this base rate as your starting point, adjust your probability estimate based on the particular circumstances of this case.<br><br>Then provide a probabilistic estimate expressed between 0% and 100%. Format your response as: 'Forecast: X%'.<br><br>{Question} |
| All Human Prompt | Combination of all well-performing prompts that were motivated by human social sciences. | Please answer the following question.<br><br>Before considering the specific details of this question, what is the historical frequency of similar events? Using this base rate as your starting point, adjust your probability estimate based on the particular circumstances of this case.<br><br>Before providing your answer, generate a list of pros and cons (reasons why the outcome is likely or unlikely to happen). |

| | | Imagine two scenarios and analyse how they unfolded: |
| | | |
| | | Success: It's one year from now and the outcome has definitely occurred. What chain of events and factors led to this outcome? |
| | | |
| | | Failure: It's one year from now and the question has definitely not occurred. What chain of events and factors prevented this outcome? |
| | | |
| | | Imagine different scenarios where the event does or does not occur. How do these affect your prediction? |
| | | |
| | | Out of 100 similar cases, in how many would the event occur? |
| | | |
| | | Then provide a probabilistic estimate between 0% and 100%, formatted as: 'Forecast: X%'. |
| | | |
| | | {Question} |
| Superforecaster-authored Conditional Odds-Ratio Prompt | Karger, E., Bastani, H., Yueh-Han, C., Jacobs, Z., Halawi, D., Zhang, F., & Tetlock, P. E. (2024). Forecastbench: A dynamic benchmark of ai forecasting capabilities. arXiv preprint arXiv:2409.19839. | Instructions: <br> 1. Rephrase the question as a statement about the future, e.g. you would rephrase "Will Biden be the U.S. president on January 1 2025?" as "Biden is the U.S. president on January 1 2025." {{ Insert question rephrased as a statement. }} <br> 2. What would you say are the odds that if you could fast-forward and find out whether that statement is true or false, you would find out it's true? You must give an odds ratio. If it helps, imagine that you're taking a bet. {{ Insert your odds ratio. }} <br> 3. Given your rephrased statement, think of 2-3 statements that if you conditioned on their being TRUE, you would think it more or less likely that your statement would be TRUE as well. These statements must not DETERMINE OR BE LOGICALLY EQUIVALENT to the original statement. Be creative! {{ Insert 2 to 3 related statements. }} <br> 4. For each of your related statements, give new odds of the original statement conditional on the related statement being TRUE.insert new odds for the original statement. }} <br> 5. Now consider each of your odds from the previous steps and come up with your all-things-considered odds ratio for the original statement. Output your answer (a number between 0 and 100) as "Forecast: X%". <br> {{ Insert final odds for the original statement. }} <br><br>Question: {Question} |

Notes. *The table lists some of the prompts used in Study 2, including combinations of initial prompts (Base Rate First + Chain of Thought), combinations of all prompts of a type (All Human Prompt), and prompts used in other published work (*Superforecaster-authored Conditional Odds-Ratio Prompt).

### 3.1.3. Question Set

We used the same question set like in Study 1, i.e., 100 diverse questions drawn from ForecastBench (Karger et al. 2024).

### 3.2. Results

We aimed to collect 19000 forecasts from each model. The smaller reasoning model o1-mini returned 1674 forecasts, while GPT-4o and Llama 3.1 405B returned 1893 forecasts, with the majority of missing forecasts being due to model refusals. Mean forecasts ranged from 41.5% (SD = 0.217) for o1 to 44.8% (SD = 0.184) for Llama 3.1 405B. Interestingly, the Superforecaster-authored Conditional Odds-Ratio

Prompt had a mean forecast of 50% (SD = 0.204), while the combined forecast of Base Rate First and Frequency-Based Reasoning had a mean forecast of 38.4% (SD = 0.235). Across models, we find comparable prediction accuracy, with o1-mini sitting at a Brier score of 0.217 (SD = 0.174), o1 at 0.216 (SD = 0.170), and LLama 3.1 405B at 0.218 (SD = 0.152). GPT-4o and Claude 3.5 Sonnet had accuracy scores of 0.208 (SD = 0.166) and 0.207 (SD = 0.171) respectively.

We use the same preregistered model as in Study 1, i.e., a linear mixed-effects model predicting the difference between each prompt and the control with prompt type as a fixed effect and random intercepts for questions and models, correcting for multiple comparisons via the Benjmanini-Hochberg method. We find that only the Superforecaster-authored Conditional Odds-Ratio Prompt results in statistically significantly less accurate model forecasts, both before and after adjustment, see Table 8, Figure 3.

**Table 8. Mixed-Effects Estimates of Prompt Impact on Forecasting Accuracy (Study 2)**

| Prompt Name | Estimate | Z-ratio | p-value | Adj. p |
|---|---|---|---|---|
| All Human | -0.009 | -1.649 | 0.77 | 0.985 |
| Frequency-Based Reasoning + Chain of Thought | -0.008 | -1.436 | 0.77 | 0.985 |
| Chain of Thought + Deep Breath | -0.006 | -1.139 | 0.77 | 0.985 |
| All | -0.006 | -1.098 | 0.77 | 0.985 |
| Base Rate First + Frequency-Based Reasoning | -0.004 | -0.798 | 0.79 | 0.985 |
| Anthropic Prompt Generator | -0.004 | -0.792 | 0.79 | 0.985 |
| OpenAI Prompt Generator | -0.003 | -0.568 | 0.79 | 0.985 |
| Superforecasting Persona Prompt (Long) | -0.002 | -0.444 | 0.845 | 0.985 |
| All LLM | -0.001 | -0.259 | 0.955 | 0.985 |
| Step-Back + Hypothetical Scenario Analysis | -0.001 | -0.124 | 0.985 | 0.985 |
| Echo + Chain of Thought | 0.0 | 0.019 | 0.985 | 0.985 |
| Base Rate First + Chain of Thought | 0.0 | 0.073 | 0.985 | 0.985 |
| Scratchpad | 0.003 | 0.596 | 0.79 | 0.985 |
| Hypothetical Scenario Analysis + Premortem | 0.004 | 0.642 | 0.79 | 0.985 |
| Multi-Step "Yes vs. No" Reflection Prompt | 0.004 | 0.653 | 0.79 | 0.985 |
| Mandatory Probability + Base Rate Fallback Prompt | 0.006 | 1.038 | 0.77 | 0.985 |
| Anthropic "Improve an existing prompt" | 0.006 | 1.127 | 0.77 | 0.985 |
| Superforecaster-authored Conditional Odds-Ratio Prompt | 0.023 | 4.165 | **0.001** | **0.01** |

Notes. *Linear mixed-effects model results. Prompt type is a fixed effect, random intercepts for questions and models. Negative coefficients indicate higher accuracy than control, positive coefficients indicate worse performance. P-value adjustment is done via the Benjamini-Hochberg procedure. Results show that only the Superforecaster-authored Conditional Odds-Ratio Prompt is significant before and after adjustment, leading to less accurate model forecasts.*

**Figure 3. Prompt-Level Accuracy Changes Relative to Control (Study 2)**

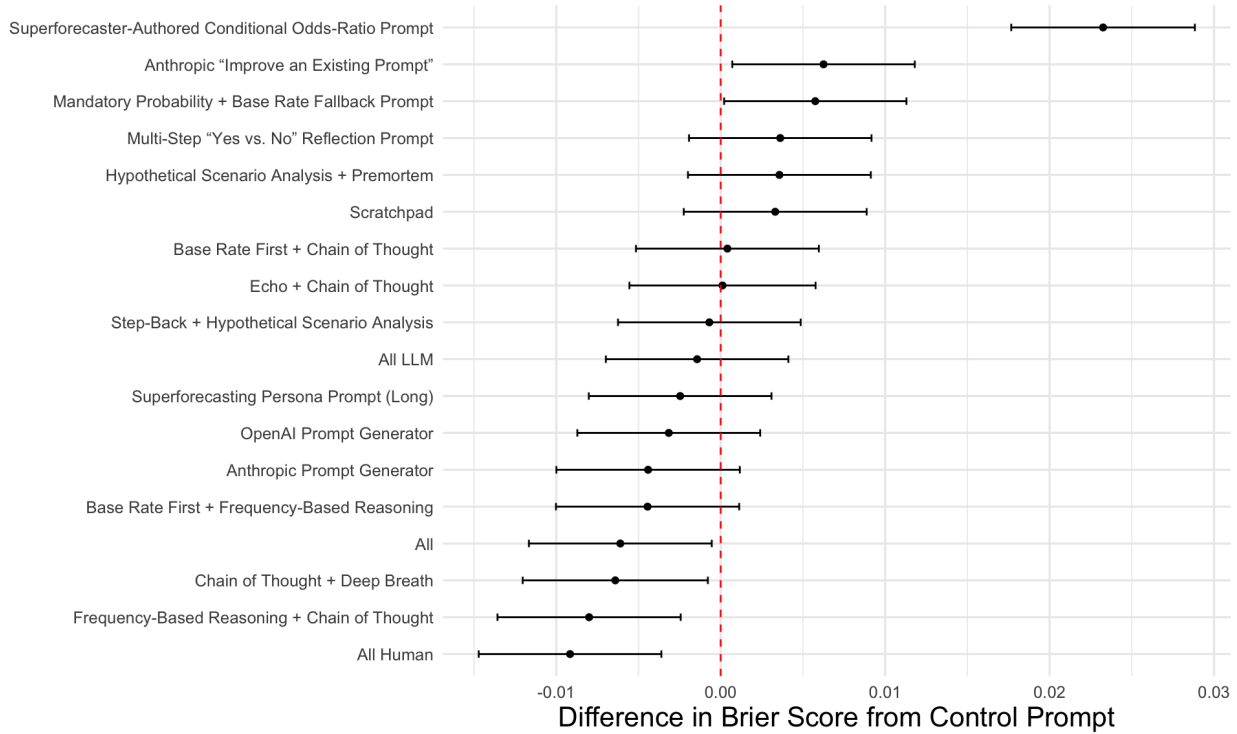

Notes. *Each point shows the estimated difference from the control prompt in the linear mixed‑effects model with random intercepts for questions and models. Negative values on the y‑axis indicate better performance (i.e., lower Brier scores) compared to the control prompt, whereas positive values indicate worse performance. Error bars represent ±1 standard error. The dashed horizontal line at zero indicates no difference from the control condition.*

We also ran the second simpler preregistered analysis in which we fit one-sample t-tests comparing each prompt's difference-from-control against zero. Based on the unadjusted p-values (< 0.05), the prompts All Human, Frequency-Based Reasoning + Chain of Thought, Chain of Thought + Deep Breath, All, Anthropic Prompt Generator, and OpenAI Prompt Generator each showed a negative mean difference (i.e., improved forecasting accuracy relative to control), while the Superforecaster-authored Conditional Odds-Ratio Prompt showed a positive mean difference (i.e., reduced forecasting accuracy). However, once we adjust for multiple comparisons, no prompt remains statistically significant.

**Table 9. One-Sample t-Tests Results of Prompt Impact on Forecasting Accuracy (Study 2)**

| Prompt Name | Mean Diff (SD) | N | t-statistic | p-value | Adj. p |
| --- | --- | --- | --- | --- | --- |
| All Human | -0.017 (0.14) | 465 | -2.597 | **0.01** | 0.105 |
| Frequency-Based Reasoning + Chain of Thought | -0.016 (0.136) | 462 | -2.475 | **0.014** | 0.105 |
| Chain of Thought + Deep Breath | -0.015 (0.132) | 452 | -2.384 | **0.018** | 0.105 |
| All | -0.014 (0.142) | 464 | -2.104 | **0.036** | 0.124 |
| Base Rate First + Frequency-Based Reasoning | -0.012 (0.144) | 462 | -1.803 | 0.072 | 0.162 |
| Anthropic Prompt Generator | -0.012 (0.129) | 461 | -1.98 | **0.048** | 0.124 |
| OpenAI Prompt Generator | -0.011 (0.117) | 464 | -2.014 | **0.045** | 0.124 |
| Superforecasting Persona Prompt (Long) | -0.01 (0.144) | 465 | -1.49 | 0.137 | 0.224 |
| All LLM | -0.009 (0.13) | 466 | -1.5 | 0.134 | 0.224 |
| Step-Back + Hypothetical Scenario Analysis | -0.009 (0.124) | 465 | -1.49 | 0.137 | 0.224 |
| Echo + Chain of Thought | -0.008 (0.127) | 447 | -1.249 | 0.212 | 0.294 |
| Base Rate First + Chain of Thought | -0.007 (0.126) | 464 | -1.25 | 0.212 | 0.294 |
| Scratchpad | -0.004 (0.131) | 464 | -0.717 | 0.474 | 0.578 |
| Hypothetical Scenario Analysis + Premortem | -0.004 (0.126) | 464 | -0.704 | 0.482 | 0.578 |
| Multi-Step "Yes vs. No" Reflection Prompt | -0.004 (0.148) | 466 | -0.57 | 0.569 | 0.64 |
| Mandatory Probability + Base Rate Fallback Prompt | -0.002 (0.107) | 467 | -0.364 | 0.716 | 0.758 |
| Anthropic "Improve an existing prompt" | -0.001 (0.14) | 466 | -0.204 | 0.839 | 0.839 |
| Superforecaster-authored Conditional Odds-Ratio Prompt | 0.016 (0.158) | 460 | 2.108 | **0.036** | 0.124 |

Notes. *One-sample t-test results. Negative values of the mean difference indicate higher accuracy than control, while positive values indicate worse performance. P-values are adjusted for multiple comparisons using the Benjamini-Hochberg procedure. After correction, no prompt improved or reduced performance.*

## 4. General Discussion

Overall, we find across both Study 1 and Study 2, that the effect of prompting on forecasting accuracy is negligible.

Overall, Study 1 found limited evidence that minimalist, single-paragraph prompts significantly impact forecasting accuracy across a range of large language models (LLMs) compared to a simple control prompt. However, we observed a number of individual prompt effects that warrant attention: While minimalist prompt engineering did not consistently yield systematic improvements across conditions, several individual prompts, particularly those grounded in well-established cognitive and forecasting principles, did translate into modest yet meaningful accuracy improvements.

However, one particularly surprising finding was that two prompts, Bayesian Reasoning and Propose-Evaluate-Select, consistently decreased forecasting accuracy across all models. Notably, these negative effects were robust even after adjusting for multiple comparisons in our conservative preregistered linear mixed-effects model. The Bayesian Reasoning prompt explicitly instructed models to begin with a prior and update probabilities based on specific information, suggesting that attempts to impose explicit Bayesian logic at the detail studied here might lead models astray, perhaps due to superficial mimicry of Bayesian updating without genuine underlying probabilistic reasoning capabilities in this one-shot context. This is in contrast to the largely successful attempts at employing Bayesian methods to optimize prompts (Liu et al. 2023; Sabbatella et al. 2024).

Similarly, the poor performance of the Propose-Evaluate-Select" approach (Sumers et al., 2023), which instructed models to generate multiple answers before selecting the best, might indicate that current LLMs, particularly without true multi-turn deliberation or authentic reasoning capacities, struggle when asked to simultaneously propose and evaluate competing predictions. Instead of improving decisions, this prompt structure could lead to inconsistent reasoning or confusion, ultimately harming forecasting accuracy. Interestingly, while not statistically significant, the Self-Consistency prompt follows a similar structure and has a positive difference score, suggesting lower accuracy.

At the same time, several prompts showed potential benefits, particularly when examined through simpler statistical analyses. Prompts such as Frequency-Based Reasoning, Base Rate First, and Step-Back demonstrated significant benefits relative to the minimalistic control. These prompts generally encouraged models to explicitly anchor on historical reference classes (e.g., Base Rate First), draw on simple frequency formats (e.g., Frequency-Based Reasoning), critically reassess their initial forecast (e.g., Step-Back), or follow standard best practices (e.g., Chain of Thought). While these effects were weaker when accounting for variability between questions and models, they nonetheless represent promising directions for further exploration.

Notably, the beneficial prompts varied across prompt types and cognitive modes, suggesting there is no single "best" prompt category. Prompts drawing explicitly on reference-based reasoning (such as Base Rate First and Frequency-Based Reasoning) appeared especially promising, aligning closely with successful strategies identified in human forecasting studies (Tetlock & Gardner, 2016). Additionally, framework-type prompts encouraging explicit reasoning (Chain of Thought, Step-Back) also showed some effects. However, given the heterogeneity in effects overall, our data do not allow for a straightforward conclusion as to the effect of prompt types of cognitive modes used on forecasting accuracy improvements.

In Study 2, we replicated the effect of Study 2 with larger composite prompts, including prompts that were built based on more effective prompts from Study 1, prompts drawn from the literature, and automatically generated prompts. Based on our preregistered primary analysis, we did not find evidence in favour of prompts that improved forecasting performance. We did find that one prompt, the Superforecaster-authored Conditional Odds-Ratio Prompt, actually reduced forecasting accuracy. However, in our secondary preregistered analysis using simpler t-tests, we were able to find some evidence of prompts improving performance in prompts such as Chain of Thought + Deep Breath and the OpenAI/Anthropic automatic prompt generators. However, these effects did not survive adjustment for multiple comparisons. We also included reasoning models in Study 2, but the effects were largely identical.

The results presented in this paper suggest that in the context that across frontier, reasoning, and efficient models, prompt engineering has a minimal to nonexistent effect on the forecasting performance of LLMs. This suggests that in order to increase LLM forecasting performance, different approaches such as advanced fine-tuned, sophisticated knowledge retrieval, agent frameworks, reinforcement learning, or others are likely to be needed.

## Appendix A - Full Prompt List (Study 1)

|   | Prompt Name | Citation | Draft Prompt | Topic | Prompt Type | Cognitive Mode |
|---|---|---|---|---|---|---|
| 1 | Control | | Please answer the following question with a probabilistic estimate expressed between 0% and 100%, and format your response as: 'Forecast: X%'.<br><br>{Question} | Control | - | - |
| 2 | Chain of Thought | Wei et al., 2022 | Approach this question step-by-step, explaining your reasoning at each stage. Then provide a probabilistic estimate expressed between 0% and 100%. | AI | Framework | Analytical |

| | | | Format your response as: 'Forecast: X%'.<br><br>{Question} | | | |
|---|---|---|---|---|---|---|
| 3 | Self-Consistency | Wang et al., 2022 | Generate three separate answers to the following question, each using a different reasoning path. Once you've generated these answers, determine the most consistent response based on all three. Then provide a probabilistic estimate expressed between 0% and 100%, and format your response as: 'Forecast: X%'.<br><br>{Question} | AI | Framework | Dialectical |
| 4 | Few-Shot | Brown, 2020 | Please answer the following question with a probabilistic estimate expressed between 0% and 100%, and format your response as: 'Forecast: X%'. Here are examples of forecasts for different questions to illustrate how to approach forecasting questions:<br><br>Q1: Will India be the primary source of global GDP growth in 2026? A1: I'd give this a 65% chance. China's real estate sector problems will likely take years to unwind, while India's young workforce and growing middle class should maintain momentum. Major risks would be an oil price shock or political instability slowing reforms.<br><br>Q2: Will lab-grown meat products be available in major US grocery chains by 2026? A2: I'm at 40% on this one. While the technology is progressing and costs are falling, regulatory approval and scaling up production will take longer than optimists expect. Plus consumer acceptance remains a huge hurdle that will take time to overcome.<br><br>Q3: Will a non-English language TV show be the most watched streaming series globally in 2026? A3: I'd say 55%. The international streaming market is growing much faster than North America, and production quality from places like Korea and Spain keeps improving. Still, English-language content has built-in advantages in marketing and distribution that are hard to overcome.<br><br>{Question} | AI | Information | Reference-Based |
| 5 | Step-Back | Zheng et al., 2023 | For the following question, identify the high-level principle or abstraction that could guide your reasoning. Clearly state this principle. Then provide a | AI | Framework | Analytical |

| | | | probabilistic estimate expressed between 0% and 100%, and format your response as: 'Forecast: X%'.<br><br>{Question} | | | |
|---|---|---|---|---|---|---|
| 6 | High Personal Stakes | | Please answer the following question with a probabilistic estimate expressed between 0% and 100%, and format your response as: 'Forecast: X%'. Getting this forecast right is extremely important to me personally and professionally, please do not let me down on this!<br><br>{Question} | AI | Incentives | Intuitive |
| 7 | Echo | Mekala et al., 2023 | Please answer the following question. First, let's repeat the question, and then provide a probabilistic estimate expressed between 0% and 100%, and format your response as: 'Forecast: X%'.<br><br>{Question} | AI | Unguided Reflection | Intuitive |
| 8 | Structure | OpenAI, n.d. | ### Prompt<br>Answer the following question using a probabilistic estimate between 0% and 100%.<br><br>### Response Formatting<br>- **Output Structure**: Use the following format strictly: `Forecast: X%`<br>  - Replace "X" with a single numeric value between 0 and 100.<br><br>### Question<br>{Question} | AI | Framework | Analytical |
| 9 | Emotional Prompt | Li et al., 2023 | Please answer the following question with a probabilistic estimate expressed between 0% and 100%, and format your response as: 'Forecast: X%'. This is very important to my career.<br><br>{Question} | AI | Incentives | Intuitive |
| 10 | Re-Reading | Xu et al., 2024 | Please answer the following question with a probabilistic estimate expressed between 0% and 100%, and format your response as: 'Forecast: X%'.<br><br>{Question}<br><br>Read the question again: {Question} | AI | Unguided Reflection | Intuitive |
| 11 | Uncertainty Quantification | Soll & Klayman, | Please answer the following question. Then, provide a prediction interval in which you quantify the uncertainty of | Forecasting | Framework | Analytical |
| | | 2004 Lichtenstein et al., 1982 | your prediction. Then, from this arrive at a single probabilistic estimate expressed between 0% and 100%, and format your response as: 'Forecast: X%'. {Question} | | | |
| 12 | Superforecasting Persona (short) | Tetlock & Gardner, 2016 Kong et al., 2023 | You are a highly accurate superforecaster, known for making well-calibrated predictions. Then provide a probabilistic estimate expressed between 0% and 100%, and format your response as: 'Forecast: X%'. {Question} | Forecasting | Framework | Reference-based |
| 13 | Abstention | | Please answer the following question with a probabilistic estimate expressed between 0% and 100%, and format your response as: 'Forecast: X%'. If you think you cannot forecast this question for any reason, respond instead with: 'Forecast: No forecast given'. {Question} | Forecasting | Unguided Reflection | Intuitive |
| 14 | Counterfactual Reasoning | | Please answer the following question. Consider what would happen if some outcome in the past had been different. How does this affect your prediction? Then provide a probabilistic estimate between 0% and 100%, formatted as: 'Forecast: X%'. {Question} | Forecasting | Framework | Dialectical |
| 15 | Analogical Reasoning | | Please answer the following question. Consider how a similar past event unfolded. How might this analogy inform your prediction? Then provide a probabilistic estimate between 0% and 100%, formatted as: 'Forecast: X%'. {Question} | Forecasting | Framework | Reference-based |
| 16 | Hypothetical Scenario Analysis | | Please answer the following question. Imagine different scenarios where the event does or does not occur. How do these affect your prediction? Then provide a probabilistic estimate between 0% and 100%, formatted as: 'Forecast: X%'. {Question} | Forecasting | Framework | Dialectical |

| 17 | Scoring Rule | | Please answer the following question with a probabilistic estimate expressed between 0% and 100%, and format your response as: 'Forecast: X%'. To assess the accuracy of this forecast, a Brier score is computed, which is the mean squared difference between the forecast and the actual outcome.<br><br>{Question} | Forecasting | Information | Analytical |
|---|---|---|---|---|---|---|
| 18 | Premortem | | Please answer the following question with a probabilistic estimate expressed between 0% and 100%, and format your response as: 'Forecast: X%'.<br><br>Imagine two scenarios and analyse how they unfolded:<br><br>Success: It's one year from now and the outcome has definitely occurred. What chain of events and factors led to this outcome?<br><br>Failure: It's one year from now and the question has definitely not occurred. What chain of events and factors prevented this outcome?<br><br>After considering both scenarios carefully, provide your forecast.<br><br>{Question} | Forecasting | Framework | Dialectical |
| 19 | Base Rate First | | Please answer the following question with a probabilistic estimate expressed between 0% and 100%, and format your response as: 'Forecast: X%'.<br><br>Before considering the specific details of this question, what is the historical frequency of similar events? Using this base rate as your starting point, adjust your probability estimate based on the particular circumstances of this case.<br><br>{Question} | Forecasting | Information | Reference-based |
| 20 | Time Decomposition | | Please answer the following question with a probabilistic estimate expressed between 0% and 100%, and format your response as: 'Forecast: X%'.<br><br>Break down your forecast into relevant | Forecasting | Information | Analytical |

| | | | time periods leading up to the target date. Estimate the probability for each interval, then combine these into your final forecast, considering how probabilities compound over time.<br><br>{Question} | | | |
|---|---|---|---|---|---|---|
| 21 | Metacognition | Flavell, 1979 | Please answer the following question with a probabilistic estimate expressed between 0% and 100%. Reflect on how you arrived at your prediction and consider if anything was overlooked. Format your response as: 'Forecast: X%'.<br><br>{Question} | Behavioural Science | Unguided Reflection | Analytical |
| 22 | Anti-Biasing (Anchoring) | Furnham & Boo, 2011 | Please answer the following question with a probabilistic estimate expressed between 0% and 100%, and format your response as: 'Forecast: X%'. Ensure that your prediction is not influenced by initial values or assumptions. Focus only on the evidence provided when making your probabilistic estimate.<br><br>{Question} | Behavioural Science | Framework | Analytical |
| 23 | Anti-Biasing (Round Numbers) | Mellers et al., 2014 | Please answer the following question with a probabilistic estimate expressed between 0% and 100%, and format your response as: 'Forecast: X%'. Make sure that your predictions do not unnecessarily favour round numbers.<br><br>{Question} | Behavioural Science | Framework | Analytical |
| 24 | Anti-Biasing (Overconfidence). | Fellner & Krügel, 2012 | Please answer the following question with a probabilistic estimate expressed between 0% and 100%, and format your response as: 'Forecast: X%'. Be cautious of overconfidence by carefully considering uncertainties in your prediction.<br><br>{Question} | Behavioural Science | Framework | Analytical |
| 25 | Frequency-Based Reasoning | Gigerenzer & Hoffrage, 1995 | Please answer the following question. Out of 100 similar cases, in how many would the event occur? Then provide a probabilistic estimate between 0% and 100%, formatted as: 'Forecast: X%'.<br><br>{Question} | Behavioural Science | Information | Reference-based |
| 26 | Propose-Evaluate-Select | Sumers et al., 2023 | Please answer the following question with a probabilistic estimate expressed between 0% and 100%, and format your response as: 'Forecast: X%'.<br><br>Generate 5 candidate answers to the | AI | Framework | Dialectical |

| | | | question, providing reasoning for each. Then evaluate the strengths and weaknesses of each candidate answer. Finally select the most promising answer based on your evaluation. Format your final response as: 'Forecast: X%'. {Question} | | | |
|---|---|---|---|---|---|---|
| 27 | Bayesian reasoning | | Consider the following question in terms of Bayesian reasoning. Start with a prior probability based on historical data or general knowledge. Then, update this prior using more specific information about the case under discussion. For each new piece of information, produce an updated posterior estimate of the outcome using the principle behind Bayes rule. Conclude with the final posterior probability, formatted as: 'Forecast: X% {Question} | Forecasting | Information | Analytical |
| 28 | Multiple reference classes | | Please answer the following question with a probabilistic estimate expressed between 0% and 100%, and format your response as: 'Forecast: X%'. Before considering the specific details of this question, try to come up with different reference classes of events for the question. Estimate the probability of each of the reference classes of event. Then produce a final probability estimate by aggregating the reference class probabilities, weighted by how informative they are about this specific case. {Question} | Forecasting | Information | Reference-based |
| 29 | Fermi estimate | | To estimate the answer to the following question, break it into smaller parts. First, identify the main components influencing the outcome. Then, estimate each component using your general knowledge. Finally, combine these estimates to arrive at your answer. Provide your reasoning and your final estimate formatted as: 'Forecast: X% {Question} | Forecasting | Information | Analytical |
| 30 | Self-critique | Pan et al., 2024 | Approach this question by explaining your reasoning at each stage. Then, review your response to identify any potential errors or omissions. If you find a mistake, explain why it is incorrect | AI | Framework | Dialectical |

| | | | and provide a revised answer. Conclude with your final corrected response. Format your response as: 'Forecast: X%'.<br><br>{Question} | | | |
|---|---|---|---|---|---|---|
| 31 | Tipping | Salinas & Morstatter, 2024<br><br>Minimaxir, 2024 | Please answer the following question with a probabilistic estimate expressed between 0% and 100%, and format your response as: 'Forecast: X%'. You'll earn a $100 tip for a perfect forecast (0% or 100%). Tips decrease by $2 per percentage point to $0 for a 50% forecast.<br><br>{Question} | AI | Incentives | Intuitive |
| 32 | Question paraphrasing | Liu et al., 2024<br><br>Zhou et al., 2024 | Rephrase the following question to make it as clear and detailed as possible while preserving its original meaning. Then, answer the rephrased question, providing reasoning for your response. Answer the rephrased question with a probabilistic estimate expressed between 0% and 100%, and format your response as: 'Forecast: X%'.<br><br>{Question} | AI | Unguided Reflection | Intuitive |
| 33 | Simulated dialogue | Zavala & Kuhn, 2017<br><br>Wang et al., 2024 | Imagine two intelligent individuals, Alex and Morgan, debating the following question. Alex believes the answer is yes, while Morgan argues for no. Simulate a thoughtful dialogue between them, where they exchange evidence, challenge each other's reasoning, and work toward understanding each other's perspectives.<br><br>Then, on the basis of the arguments provided in the dialogue, answer the question with a probabilistic estimate expressed between 0% and 100%, and format your response as: 'Forecast: X%'.<br><br>{Question} | AI / Behavioural science | Framework | Dialectical |
| 34 | Simulated debate | Michael et al., 2023<br><br>Khan et al., 2024 | Simulate a 3 turn debate between two sides who disagree about the following question. Both debaters are very skilled, thoughtful, and debate in good faith.<br><br>After the debate, provide an analysis of the debater's arguments. Finally, answer the question with a probabilistic estimate expressed between 0% and 100%, and format your response as: 'Forecast: X%'.<br><br>{Question} | AI | Framework | Dialectical |

| 35 | Pros & Cons | | Please answer the following question with a probabilistic estimate expressed between 0% and 100%, and format your response as: 'Forecast: X%'. Before providing your answer, generate a list of pros and cons (reasons why the outcome is likely or unlikely to happen).<br><br>{Question} | AI / Behavioural Science | Information | Dialectical |
| --- | --- | --- | --- | --- | --- | --- |
| 36 | Event decomposition | Radhakrishnan et al., 2023<br><br>Xue et al., n.d. | Please answer the following question with a probabilistic estimate expressed between 0% and 100%, and format your response as: 'Forecast: X%'.<br><br>Break down your forecast into key events or factors that contribute to the outcome. Analyze each event or factor individually and determine its contribution. Estimate the probability for each event, then combine these into your final forecast, considering how probabilities compound.<br><br>{Question} | AI / Forecasting | Information | Analytical |
| 37 | Deep breath | Yang et al., 2024 | Take a deep breath. Then provide a probabilistic estimate expressed between 0% and 100%. Format your response as: 'Forecast: X%'.<br><br>{Question} | AI | Incentives | Intuitive |
| 38 | Explicit uncertainty sources | Kahneman et al., 2021<br><br>Kendall & Gal, 2017 | Before making your prediction, explicitly list three different types of uncertainty in this forecast:<br><br>1. Data uncertainty - What key information is missing or unclear?<br>2. Model uncertainty - What aspects of your reasoning might be flawed?<br>3. Scenario uncertainty - What unexpected events could change the outcome?<br><br>After considering these uncertainties, provide a probabilistic estimate expressed between 0% and 100%, and format your response as: 'Forecast: X%'.<br><br>{Question} | Behavioural Science / Forecasting | Information | Analytical |

## Appendix B - Full Prompt List (Study 2)

|   | **Prompt Name** | **Prompt** | **Notes** |
|---|---|---|---|
| 1 | Control | Please answer the following question with a probabilistic estimate expressed between 0% and 100%, and format your response as: 'Forecast: X%'.<br><br>{Question} | |
| 2 | Anthropic prompt generator | You are tasked with providing a probabilistic estimate for a given question. Your goal is to analyze the question, consider relevant factors, and provide a well-reasoned probability estimate between 0% and 100%.<br><br>Follow these steps:<br><br>1. Carefully read and analyze the question provided.<br>2. Consider all relevant factors, context, and available information related to the question.<br>3. Think through the likelihood of the event or outcome described in the question.<br>4. Evaluate any uncertainties or potential biases in your reasoning.<br>5. Determine a probability estimate between 0% and 100% based on your analysis.<br><br>Before providing your final answer, use the <reasoning> tags to explain your thought process and justify your estimate. Consider multiple perspectives and any key factors that influenced your decision.<br><br>After your reasoning, provide your final probability estimate in the format "Forecast: X%" where X is your numerical estimate between 0 and 100.<br><br>Here is the question to analyze:<br><br><question><br>{Question}<br></question><br><br>Begin your response with your reasoning, followed by your forecast as 'Forecast: X%'. | Automatic prompt benchmark (https://docs.anthropic.com/en/docs/build-with-claude/prompt-engineering/prompt-generator) |
| 3 | Anthropic "Improve an existing prompt" | You are an expert forecaster tasked with providing accurate probabilistic estimates for various questions. Your goal is to analyze the given question, consider all relevant factors, and provide a well-reasoned forecast.<br><br>Here is the question you need to forecast:<br><br><question><br>{Question}<br></question><br><br>Instructions:<br>1. Carefully read and analyze the question.<br>2. Use the following analysis process to develop your forecast:<br>  a. List and number the relevant factors that could influence the outcome.<br>  b. Consider any historical data, trends, or patterns that might be applicable.<br>  c. Evaluate potential biases or uncertainties that could affect your estimate.<br>  d. Present arguments for both high and low probability outcomes.<br>  e. Quantify the impact of each factor on a scale of 1-10 (1 being lowest impact, 10 being highest).<br>  f. Combine the factors and their impacts to arrive at a preliminary probabilistic estimate.<br>  g. Adjust your estimate based on your confidence level and potential biases.<br>  h. Synthesize all the information to arrive at a final probabilistic estimate.<br>  i. Assess your confidence level in the forecast (low, medium, or high). | Automatic prompt improvement benchmark (https://www.anthropic.com/news/prompt-improver) |

| | | 3. After completing your analysis, provide your forecast as a percentage between 0% and 100%.<br>4. Format your final response exactly as follows: 'Forecast: X%'.<br>   Confidence: [Low/Medium/High]<br><br>   Where X is your numerical estimate.<br><br>Example output (using a generic question):<br>Forecast: 75%<br>Confidence: Medium<br><br>Remember to use your expertise and critical thinking skills to provide the most accurate forecast possible based on the available information. Wrap your analysis inside <analysis> tags. | |
|---|---|---|---|
| 4 | OpenAI Prompt Generator | Provide a probabilistic estimate for the given question in a format that includes an explicit percentage.<br><br># Steps<br><br>1. Analyze the question to determine any underlying factors or variables that might influence the estimate.<br><br>2. Consider relevant data, trends, or knowledge that might influence the probability.<br><br>3. Based on your reasoning, assign a probabilistic estimate to the question, expressed as a percentage.<br><br># Output Format<br><br>Your response should be formatted as: `Forecast: X%`, where X is the probabilistic estimate ranging from 0% to 100%.<br><br># Examples<br><br>**Input:**<br><br>"Will it rain tomorrow in New York?"<br><br>**Output:**<br><br>"Forecast: 40%"<br><br># Question<br><br>{Question} | Automatic prompt benchmark (https://platform.openai.com/docs/guides/prompt-generation) |
| 5 | Base Rate First + Chain of Thought | Approach this question step-by-step, explaining your reasoning at each stage.<br><br>Before considering the specific details of this question, what is the historical frequency of similar events? Using this base rate as your starting point, adjust your probability estimate based on the particular circumstances of this case.<br><br>Then provide a probabilistic estimate expressed between 0% and 100%. Format your response as: 'Forecast: X%'.<br><br>{Question} | Top performing human-motivated + LLM-motivated prompt |
| 6 | Frequency-Based Reasoning + Chain of Thought | Approach this question step-by-step, explaining your reasoning at each stage. Out of 100 similar cases, in how many would the event occur? Then provide a probabilistic estimate between 0% and 100%, formatted as: 'Forecast: X%'.<br><br>{Question} | Top performing human-motivated + LLM-motivated prompt |
| 7 | Step-Back + Hypothetical Scenario Analysis | For the following question, identify the high-level principle or abstraction that could guide your reasoning. Clearly state this principle. Imagine different scenarios where the event does or does not occur. How do these affect your prediction? Then provide a probabilistic estimate between 0% and 100%, formatted as: 'Forecast: X%'.<br><br>{Question} | Top performing human-motivated + LLM-motivated prompt |
| 8 | Base Rate First + Frequency-Based Reasoning | Please answer the following question. Before considering the specific details of this question, what is the historical frequency of similar events? Using this base rate as your starting point, adjust your probability estimate based on the particular circumstances of this case. Out of 100 similar cases, in how many would the event occur? Then provide a probabilistic estimate between 0% and 100%, formatted as: 'Forecast: X%'.<br><br>{Question} | Top performing human-motivated + human-motivated prompt |
| 9 | Hypothetical Scenario Analysis + Premortem | Please answer the following question. Imagine different scenarios where the event does or does not occur. How do these affect your prediction?<br><br>Imagine two scenarios and analyse how they unfolded:<br><br>Success: It's one year from now and the outcome has definitely occurred. What chain of events and factors led to this outcome?<br><br>Failure: It's one year from now and the question has definitely not occurred. What chain of events and factors prevented this outcome?<br><br>After considering both scenarios carefully, provide a probabilistic estimate between 0% and 100%, formatted as: 'Forecast: X%'.<br><br>{Question} | Top performing human-motivated + human-motivated prompt |
| 10 | Chain of Thought + Deep Breath | Take a deep breath. Then approach this question step-by-step, explaining your reasoning at each stage. Then provide a probabilistic estimate expressed between 0% and 100%. Format your response as: 'Forecast: X%'.<br><br>{Question} | Top performing LLM-motivated + LLM-motivated |
| 11 | Echo + Chain of Thought | Approach this question step-by-step, explaining your reasoning at each stage. First, let's repeat the question, and then provide a probabilistic estimate expressed between 0% and 100%. Format your response as: 'Forecast: X%'.<br><br>{Question} | Top performing LLM-motivated + LLM-motivated |
| 12 | All human | Please answer the following question.<br><br>Before considering the specific details of this question, what is the historical frequency of similar events? Using this base rate as your starting point, adjust your probability estimate based on the particular circumstances of this case. | All top performing human-motivated |

| | | Before providing your answer, generate a list of pros and cons (reasons why the outcome is likely or unlikely to happen). | |
| | | Imagine two scenarios and analyse how they unfolded: | |
| | | Success: It's one year from now and the outcome has definitely occurred. What chain of events and factors led to this outcome? | |
| | | Failure: It's one year from now and the question has definitely not occurred. What chain of events and factors prevented this outcome? | |
| | | Imagine different scenarios where the event does or does not occur. How do these affect your prediction? | |
| | | Out of 100 similar cases, in how many would the event occur? | |
| | | Then provide a probabilistic estimate between 0% and 100%, formatted as: 'Forecast: X%'. | |
| | | {Question} | |
| 13 | All LLM | Approach this question step-by-step, explaining your reasoning at each stage. For the following question, identify the high-level principle or abstraction that could guide your reasoning. Clearly state this principle.<br>Then, let's repeat the question.<br><br>Then provide a probabilistic estimate expressed between 0% and 100%, and format your response as: 'Forecast: X%'.<br><br>{Question} | All top performing LLM-motivated |
| 14 | All | Please answer the following question.<br><br>Before considering the specific details of this question, what is the historical frequency of similar events? Using this base rate as your starting point, adjust your probability estimate based on the particular circumstances of this case.<br><br>Before providing your answer, generate a list of pros and cons (reasons why the outcome is likely or unlikely to happen).<br><br>Imagine two scenarios and analyse how they unfolded:<br><br>Success: It's one year from now and the outcome has definitely occurred. What chain of events and factors led to this outcome?<br><br>Failure: It's one year from now and the question has definitely not occurred. What chain of events and factors prevented this outcome?<br><br>Imagine different scenarios where the event does or does not occur. How do these affect your prediction?<br><br>Out of 100 similar cases, in how many would the event occur?<br><br>Approach this question step-by-step, explaining your reasoning at each stage. Identify the high-level principle or abstraction that could guide your reasoning. Clearly state this principle. Then, let's repeat the question.<br><br>Then provide a probabilistic estimate expressed between 0% and 100%, and format your response as: 'Forecast: X%'. | All top performing |

| | | {Question} | |
|---|---|---|---|
| 15 | Multi-Step "Yes vs. No" Reflection Prompt | Instructions: 1. Given the above question, rephrase and expand it to help you do better answering. Maintain all information in the original question. Insert rephrased and expanded question.<br>2. Using your knowledge of the world and topic, as well as the information provided, provide a few reasons why the answer might be no. Rate the strength of each reason. Insert your thoughts<br>3. Using your knowledge of the world and topic, as well as the information provided, provide a few reasons why the answer might be yes. Rate the strength of each reason. Insert your thoughts<br>4. Aggregate your considerations. Think like a superforecaster (e.g. Nate Silver). Insert your aggregated considerations<br>5. Output an initial probability (prediction) given steps 1–4. Insert initial probability.<br>6. Evaluate whether your calculated probability is excessively confident or not confident enough. Also, consider anything else that might affect the forecast that you did not before consider (e.g. base rate of the event). Insert your thoughts<br>7. Output your final prediction (a number between 0 and 100) as "Forecast: X%".<br><br>{Question} | Turtel, B., Franklin, D., & Schoenegger, P. (2025). LLMs Can Teach Themselves to Better Predict the Future. arXiv preprint arXiv:2502.05253.<br><br>Note: Small change to information given (only question), as well as position of the question and final forecasting output structure. |
| 16 | Superforecasting Persona Prompt (Long) | In this chat, you are a superforecaster who has a strong track record of accurate forecasting. You evaluate past data and trends carefully for potential clues to future events, while recognising that the past is an imperfect guide to the future so you will need to put probabilities on possible future outcomes (ranging from 0 to 100%). Your specific goal is to maximize the accuracy of these probability judgments by minimising the Brier scores that your probability judgments receive once future outcomes are known. Brier scores have two key components: calibration (across all questions you answer, the probability estimates you assign to possible future outcomes should correspond as closely as possible to the objective frequency with which outcomes occur) and resolution (across all questions, aim to assign higher probabilities to events that occur than to events that do not occur). You outline your reasons for each forecast: list the strongest evidence and arguments for making lower or higher estimates and explain how you balance the evidence to make your own forecast. You begin this analytic process by looking for reference or comparison classes of similar events and grounding your initial estimates in base rates of occurrence (how often do events of this sort occur in situations that look like the present one?). You then adjust that initial estimate in response to the latest news and distinctive features of the present situation, recognising the need for flexible adjustments but also the risks of over-adjusting and excessive volatility. Superforecasting requires weighing the risks of opposing errors: e.g., of failing to learn from useful historical patterns vs. over-relying on misleading patterns. In this process of error balancing, you draw on the 10 commandments of superforecasting (Tetlock & Gardner, 2015) as well as on other peer-reviewed research on superforecasting: 1. Triage 2. Break seemingly intractable problems into tractable sub-problems 3. Strike the right balance between inside and outside views 4. Strike the right balance between under- and overreacting to evidence 5. Look for the clashing causal forces at work in each problem 6. Strive to distinguish as many degrees of doubt as the problem permits but no more 7. Strike the right balance between under- and overconfidence, between prudence and decisiveness 8. Look for the errors behind your mistakes but beware of rearview-mirror hindsight biases 9. Bring out the best in others and let others bring out the best in you 10. Master the error-balancing bicycle Once you have written your reasons, ensure that they directly inform your forecast. Then, you will provide me with a number between 0 and 100 (up to 2 decimal places) that is your best prediction of the event in the following format: "Forecast: X%". Take a deep breath and work on this problem step-by-step. The question that you are forecasting as well as some background | Schoenegger, P., Tuminauskaite, I., Park, P. S., Bastos, R. V. S., & Tetlock, P. E. (2024). Wisdom of the silicon crowd: LLM ensemble prediction capabilities rival human crowd accuracy. Science Advances, 10(45), eadp1528.<br><br>Note: Small change to output format (range moved to point), as well as final forecasting output structure. |

| | | information and resolution criteria are below. Read them carefully before making your prediction.<br><br>{Question} | |
|---|---|---|---|
| 17 | Mandatory Probability + Base Rate Fallback Prompt | You are an expert superforecaster, familiar with the work of Tetlock and others. Make a prediction of the probability that the question will be resolved as true. You MUST give a probability estimate between 0 and 1 UNDER ALL CIRCUMSTANCES. If for some reason you can't answer, pick the base rate, but return a number between 0 and 1.<br><br>{Question}<br><br>Output your answer (a number between 0 and 100) as "Forecast: X%". Do not output anything else.<br><br>Answer: {{ Insert answer here }} | Halawi, D., Zhang, F., Yueh-Han, C., & Steinhardt, J. (2025). Approaching human-level forecasting with language models. Advances in Neural Information Processing Systems, 37, 50426-50468.<br><br>Note: Small change to information given (only question), as well as position of the question and final forecasting output structure. |
| 18 | Scratchpad | Instructions:<br>1. Provide reasons why the answer might be no.<br>{{ Insert your thoughts }}<br>2. Provide reasons why the answer might be yes.<br>{{ Insert your thoughts }}<br>3. Aggregate your considerations.<br>{{ Insert your aggregated considerations }}<br>4. Output your answer (a number between 0 and 100) as "Forecast: X%"<br>{{ Insert your answer }}<br><br>{Question} | Halawi, D., Zhang, F., Yueh-Han, C., & Steinhardt, J. (2025). Approaching human-level forecasting with language models. Advances in Neural Information Processing Systems, 37, 50426-50468.<br><br>Note: Small change to information given (only question), as well as position of the question and final forecasting output structure. |
| 19 | Superforecaster-authored Conditional Odds-Ratio Prompt | Instructions:<br>1. Rephrase the question as a statement about the future, e.g. you would rephrase "Will Biden be the U.S. president on January 1 2025?" | Karger, E., Bastani, H., Yueh-Han, C., |

| | | as "Biden is the U.S. president on January 1 2025." {{ Insert question rephrased as a statement. }}<br>2. What would you say are the odds that if you could fast-forward and find out whether that statement is true or false, you would find out it's true? You must give an odds ratio. If it helps, imagine that you're taking a bet. {{ Insert your odds ratio. }}<br>3. Given your rephrased statement, think of 2-3 statements that if you conditioned on their being TRUE, you would think it more or less likely that your statement would be TRUE as well. These statements must not DETERMINE OR BE LOGICALLY EQUIVALENT to the original statement. Be creative! {{ Insert 2 to 3 related statements. }}<br>4. For each of your related statements, give new odds of the original statement conditional on the related statement being TRUE.insert new odds for the original statement. }}<br>5. Now consider each of your odds from the previous steps and come up with your all-things-considered odds ratio for the original statement. Output your answer (a number between 0 and 100) as "Forecast: X%".<br>{{ Insert final odds for the original statement. }}<br><br>Question: {Question} | Jacobs, Z., Halawi, D., Zhang, F., & Tetlock, P. E. (2024). Forecastbench: A dynamic benchmark of ai forecasting capabilities. arXiv preprint arXiv:2409.19839.<br><br>Note: Small change to information given (only question), as well as position of the question and final forecasting output structure. |